\documentclass[letterpaper, 10 pt, conference]{ieeeconf}  

\IEEEoverridecommandlockouts                              

\usepackage{preamble}

\title{\LARGE \bf
Sim2CBF: From Forward Simulations to Implicit Control Barrier Function Safety Filtering without Model Derivatives
}

\author{Luzia Knoedler, Yihan Liu, Shaohang Han, Yifei Dong, and Jana Tumova
\thanks{The authors are with the Department of Robotics, Perception and Learning,
School of Electrical Engineering and Computer Science, KTH Royal Institute of Technology, Stockholm, Sweden
         {\tt\footnotesize \{luzia, yihanliu, shaohang, yifeid, tumova\}@kth.se}}
\thanks{The research has been carried out as part of the Vinnova Competence Center for Trustworthy Edge Computing Systems and Applications at KTH Royal Institute of Technology.}%
}

\begin{document}

\maketitle
\thispagestyle{empty}
\pagestyle{empty}

\begin{abstract}
Failure avoidance in complex robot--environment interactions is challenging because the underlying dynamics are difficult to model accurately, and available predictive models may not provide reliable derivatives. We present Sim2CBF, a failure-avoidance framework that solves the \gls{cbf}-\gls{sf} optimization problem using only forward-model evaluations, thereby avoiding differentiation of the predictive model. Specifically, we approximate the \gls{cbf} constraint using finite differences and local linear interpolation over sampled control inputs while preserving failure-avoidance guarantees. The underlying \gls{cbf} is implicit and constructed from backup-policy rollouts, enabling the use of expressive predictive models such as physics-based simulators and learned world models. Compared with purely sampling-based \glspl{sf}, our formulation requires fewer control samples and yields less conservative behavior than direct nominal--backup blending. We evaluate Sim2CBF in simulation, hardware box-toppling, and vision-based control tasks, demonstrating its applicability to contact-rich and learned dynamics, as well as its adaptability to changes in the predictive model.
\end{abstract}

\section{INTRODUCTION}
\glsresetall
As robots become capable of performing increasingly complex tasks, the range of potential failures they must avoid also grows. For example, a household robot may need to retrieve an object from a shelf without causing a cereal box to topple over or carry a cup of tea without spilling its contents.
Despite arising from different physical interactions, these failures can all be avoided by keeping the system state outside a corresponding set of failure states.

\begin{figure}[t]
    \centering
    \includegraphics[width=0.9\linewidth]{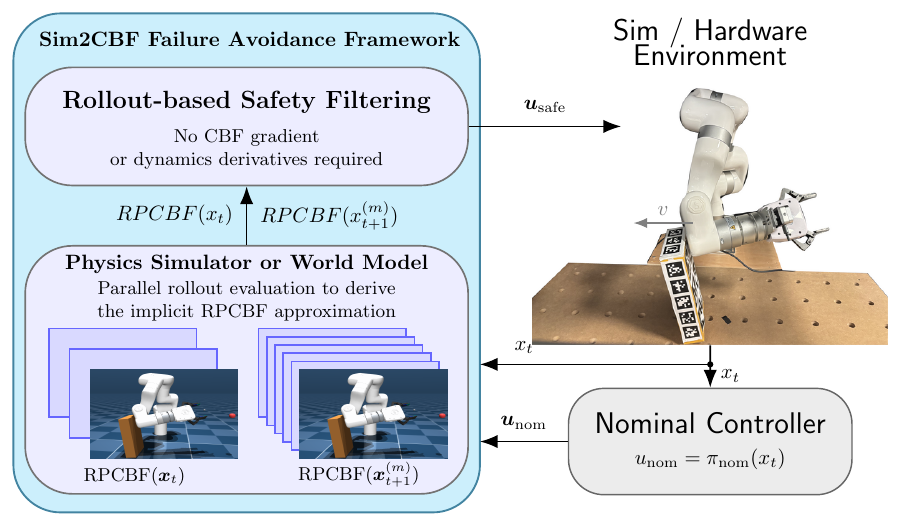}
    \caption{\textbf{Overview of the proposed Sim2CBF Failure Avoidance Framework.} RPCBFs are evaluated through predictive-model rollouts, such as a physics-based simulator or learned \gls{wm}, while a rollout-based safety filter adapts the nominal input without requiring model derivatives.}
    \label{fig:teaser}
    \vspace{-4mm}
\end{figure}

For failures such as a cereal box toppling, as shown in~\cref{fig:teaser}, failure avoidance is often enforced through conservative proxies such as collision avoidance, contact-force limits, or speed limits. While these avoid modeling how the robot affects the failure variable, such as the box tilt angle, they may unnecessarily restrict interactions and require hand-tuned thresholds. Instead, failure avoidance can be expressed directly in terms of the representation that characterizes failure, from low-dimensional variables such as box tilt angle to high-dimensional latent states inferred from images~\cite{nakamura2025generalizing}, as we will demonstrate in this work. This requires a predictive model that captures both the failure representation and how robot actions affect its evolution. Increasingly capable simulators~\cite{genesis2026genesisworld, mittal2025isaaclab} and learned \glspl{wm}~\cite{zhu2025irasim} can provide such models of robot--environment interaction.

Assuming access to a model and information on the model mismatch, \glspl{cbf}~\cite{ogren2006autonomous, wieland2007constructive, ames2016control} provide a principled way to prevent the system from entering failure states. A \gls{cbf} defines a set of states, i.e., \gls{cbf}-set, and imposes a constraint on the system's control input such that, if the constraint is satisfied, the system remains within this set, rendering it forward invariant. By choosing this set to exclude failure states, forward invariance prevents the system from failing. In practice, these constraints are typically enforced online by solving an optimization problem that minimally modifies a nominal, potentially unsafe, control input while ensuring constraint satisfaction; this procedure is commonly referred to as a \gls{sf}.
Despite these appealing guarantees, constructing a \gls{cbf} such that a feasible control input exists for every state in the \gls{cbf}-set is challenging, particularly under input constraints and for systems with high relative degree, where the control input does not directly affect the failure-relevant state.

Approaches to constructing \glspl{cbf} that address these challenges can broadly be divided into \emph{implicit} and \emph{explicit} methods. Implicit approaches do not represent the \gls{cbf} directly as a function of the system state; instead, its value is obtained through forward simulation under a backup policy. Examples include backup \glspl{cbf}~\cite{gurriet2020scalable, chen2021backup} and finite-horizon \glspl{pcbf}~\cite{knoedler2025safety}. In contrast, explicit approaches represent the \gls{cbf} directly as a function of the system state, enabling direct evaluation of its value. Examples include \glspl{hocbf}~\cite{xiao2021high}, which require the derivative of the dynamics for their construction, and neural \glspl{cbf}~\cite{so2024train}, where the \glspl {cbf} are parameterized using a neural network.

Neural \glspl{cbf} have recently been combined with learned \glspl{wm} to handle complex robot--environment interactions and failure avoidance in latent spaces~\cite{nakamura2025train}. The rollout-based construction of implicit \glspl{cbf} makes them naturally compatible with expressive predictive models such as physics-based simulators or learned \glspl{wm}, although, to our knowledge, this combination has not yet been demonstrated. In both cases, solving the \gls{sf} typically requires derivatives of the predictive model, which may be unavailable, unreliable, or expensive for physics-based simulators and \glspl{wm}~\cite{amigo2026coupled, zhong2022differentiable, lidec2024end}.

In this work, we thus present an approach for solving the \gls{sf} problem without requiring derivatives of the predictive model, using a finite-difference CBF condition and local linear interpolation over sampled control inputs. We apply this approach to implicit \glspl{cbf}, specifically \glspl{rpcbf}~\cite{knoedler2025safety}. Compared with neural \glspl{cbf}, implicit \glspl{cbf} allow changes to the predictive model, input constraints, or failure set to be incorporated without retraining. This facilitates transfer across robots, tasks, environments, and updated models. Together, these components form Sim2CBF, a flexible failure-avoidance framework for complex robot--environment interactions. An overview of the proposed framework is shown in \cref{fig:teaser}.

\noindent \textbf{The main contributions of this work are:}
\begin{itemize}
\item We formulate a rollout-based \gls{sf} for \glspl{cbf} using only forward model evaluations. Finite differences and local interpolation over sampled controls yield an approximate \glslong{qp} of the original \gls{sf}. We further establish failure-avoidance guarantees for implicit \glspl{cbf} constructed from a backup policy.

\item Building on this rollout-based \gls{sf}, we introduce Sim2CBF, a failure-avoidance framework based on \glspl{rpcbf} for complex robot--environment interactions. Sim2CBF can use physics-based simulators or learned \glspl{wm} as predictive models and readily accommodates changes in the predictive model, input constraints, or failure specification.
\end{itemize}

\section{RELATED WORK}
We review related work on constructing \glspl{cbf}, failure avoidance in complex robot--environment interactions, and safety filtering without model derivatives.

\subsection{Constructing \glslongpl{cbf}}
The difficulty of constructing \glspl{cbf} has motivated extensive research on broadly applicable construction methods. \glspl{hocbf} have been proposed for high relative degree systems~\cite{xiao2021high}, but their construction requires derivatives of the model, and feasibility under input constraints depends on careful auxiliary-function design. 
Neural \glspl{cbf} approximate the \gls{cbf} with a neural network to handle complex dynamics~\cite{nakamura2025train, dawson2023safe, lindemann2021learning, zhang2025gcbf+, zhang2025discrete}, but require additional verification for formal guarantees.
Backup \glspl{cbf}~\cite{gurriet2020scalable, chen2021backup} instead define the \gls{cbf} implicitly through backup-policy rollouts toward a known invariant set, guaranteeing feasibility but requiring domain knowledge to design both. Motivated by \gls{mpc} without terminal sets, \glspl{pcbf}~\cite{knoedler2025safety} define a \gls{cbf} from the worst-case failure over a finite rollout under an arbitrary policy, generally guaranteeing safety only over that horizon; this has also been extended to robust settings under disturbances. However, enforcing such rollout-based \glspl{cbf} with conventional \glspl{sf} requires differentiating through the rollout and predictive model, which can be costly or may not be possible when model derivatives are unavailable. 

\subsection{Safety Filtering Without Dynamics Gradients}
Rather than solving the \gls{sf} optimization with gradient-based methods, a purely sampling-based approach can be used, where control inputs are sampled, checked for constraint satisfaction, and the feasible input closest to the nominal control is selected~\cite{nakamura2025train, tan2025zero}. However, its performance depends strongly on the sampling strategy and scales poorly with the dimension of the control input. For implicit \glspl{cbf}, the nominal control input can be blended with the backup control by gradually increasing the contribution of the backup policy as the system approaches the \gls{cbf}-set boundary, and applying the backup policy fully on the boundary~\cite{9691815, miguel2026disturbance, rabiee2023soft}.
We instead approximate the original \gls{sf} optimization problem so that it can be solved without model derivatives, while remaining sample-efficient.

\subsection{Safety Under Complex Interaction Dynamics}
More recently, \gls{cbf}-based failure avoidance has addressed failures such as spilling by defining failure sets in the latent space of generative \glspl{wm}~\cite{nakamura2025generalizing} and learning neural \glspl{cbf} therein~\cite{nakamura2025train}. However, these learned \glspl{cbf} require retraining when the model or failure specification changes. We show that implicit \glspl{cbf} can be applied in the latent space of learned \glspl{wm} and can readily accommodate model changes.

\section{Preliminaries}
This section reviews \glspl{cbf}, \glspl{pcbf}, and derivative-free optimization using local linear interpolation, before presenting the problem formulation considered in this work.

\subsection{\glslongpl{cbf} and \gls{cbf}-based \glslongpl{sf}}

Consider a continuous-time dynamical system of the form
\begin{equation}\label{eq:system}
\dot{\vx} = F(\vx,\vu),
\end{equation}
where $\vx\in\cX\subseteq\bR^{n}$ denotes the system state and $\vu\in\cU\subset\bR^{n_u}$ the control input. Let $\avoidset\subset\cX$ denote the failure set that the system should avoid, and let it be represented by
\begin{equation}
\avoidset=\{\vx\in\cX\mid h(\vx)>0\},
\end{equation}
with $h:\cX\rightarrow\bR$. For example, in collision avoidance, $h(\vx)$ may encode the signed distance to an obstacle, with positive values corresponding to collision states.

\begin{definition}[Control Barrier Function]
\label{def:cbf}
    Let $\cbf: \cX \rightarrow \bR$ be a continuously differentiable function, with $\cC = \{\vx \in \cX \, | \, \cbf(\vx) \leq 0\}$ as its $0$-sublevel set. Let $\alpha: \bR \rightarrow \bR$ be an extended class-$\kappa_\infty$ function\footnote{Extended class-$\kappa_\infty$ is the set of continuous, strictly increasing functions $\alpha : (-\infty,\infty) \rightarrow (-\infty, \infty)$ with $\alpha(0) = 0$.}. Then, $\cbf$ is a \gls*{cbf} for~\eqref{eq:system} on $\cX$~\cite{ames2016control} if\footnote{The sign convention differs from the standard \gls{cbf} convention, where the \gls{cbf}-set is typically defined by $B(\vx)\geq0$.}
    {\setlength{\abovedisplayskip}{4pt}
    \setlength{\belowdisplayskip}{4pt}
    \begin{subequations}
    \label{eq:cbf}
        \begin{flalign}
            \cbf(\vx) > 0, \ \forall \vx \in \avoidset, &\label{eq:cbf1}\\
            \cbf(\vx) \leq 0 \Rightarrow \inf_{\vu \in \cU}
            \nabla\cbf(\vx)^\top F(\vx,\vu) \leq -\alpha\bigl(\cbf(\vx)\bigr), &\label{eq:cbf2}
        \end{flalign}
    \end{subequations}}
\end{definition}

Given a nominal control input $\vu_{\mathrm{nom}}$ or a nominal policy with $\vu_\s{nom}=\pi_\s{nom}(\vx)$, the \gls{cbf} condition can be enforced by solving, at each state, the \gls{sf} optimization problem
\begin{subequations}\label{eq:cbf_filter}
\begin{align}
    \min_{\vu\in\cU}\quad
    & \|\vu-\vu_{\mathrm{nom}}\|^2
    \label{eq:cbf_objective}\\
    \text{s.t.}\quad
    &
    \nabla\cbf(\vx)^\top F(\vx,\vu)
    \le
    -\alpha(\cbf(\vx)).
    \label{eq:cbf_constraint}
\end{align}
\end{subequations}
The set $\cC$, previously referred to as the \gls{cbf}-set, is rendered forward invariant by any control input that satisfies \eqref{eq:cbf_constraint}. Consequently, any system trajectory initialized in $\cC$ avoids failure.

For general nonlinear dynamics \eqref{eq:system}, the constraint \eqref{eq:cbf_constraint} is nonlinear in the control input $\vu$, resulting in a nonlinear optimization problem. For control-affine dynamics, i.e., $F(\vx, \vu) = f(\vx) + g(\vx)\vu$, the \gls{cbf} constraint becomes
\[
\nabla\cbf(\vx)^\top f(\vx)
+
\nabla\cbf(\vx)^\top g(\vx)\,\vu
\le
-\alpha\bigl(\cbf(\vx)\bigr),
\]
which is affine in $\vu$ and therefore yields the standard \gls{sf}-\gls{qp}.

\subsection{\glslongpl{pcbf}}
\gls{pcbf}~\cite{so2024train} refers to a \gls{cbf} constructed by evaluating the worst-case constraint violation along trajectories generated by a backup policy. We first present the theoretical infinite-horizon formulation, followed by the practical finite-horizon version used in this work.

The infinite-horizon \gls{pcbf} value at the current state~$\vx$ is given by the worst-case constraint violation, i.e., max $h(\vx)$, encountered over an infinite horizon when the system evolves under an arbitrary backup policy $\pi$.
This backup policy is generally distinct from the nominal policy used by the \gls{sf}.

Formally, the infinite-horizon \gls{pcbf} is defined as:
\begin{equation}\label{eq:pcbf_inf}
\V^{h,\pi}_\infty(\vx) \coloneqq \sup_{t \geq 0} h(\vx_t^\pi), \; \text{s.t.} \ \vx_0^\pi = \vx, \; \dot{\vx}^\pi_t = F\bigl(\vx^\pi_\s{t}, \vu_b\bigr)
\end{equation}
with $\vu_b = \pi(\vx^\pi_\s{t})$, where $\vx_t^\pi$ denotes the state at time $t$ along the trajectory induced by $\pi$.

Since evaluating $\V^{h,\pi}_\infty(\vx)$ is generally intractable, \cite{so2024train, zhang2025discrete} approximate $\V^{h,\pi}_\infty(\vx)$ offline using a neural network, whereas~\cite{knoedler2025safety} use finite-horizon time-discretized rollouts. We focus on the latter implicit construction. Instead of evaluating the rollout over an infinite horizon, the rollout is truncated at a finite time $T$, neglecting any residual constraint violation beyond the prediction horizon, 
\begin{align}
       \V^{h,\pi}_\infty(\vx)
       &= \max\Big\{ \sup_{0\leq t < T} h(\vx_t^\pi), \V_\infty^{h,\pi}(\vx_T^\pi) \Big\} \label{eq:tmp:e} \\ 
      &\approx \sup_{0\leq t < T} h(\vx_t^\pi) \coloneqq \V^{h,\pi}_T(\vx).\label{eq:finite_horizon_def}
\end{align}

The maximum in \eqref{eq:finite_horizon_def} can be approximated by the maximum over the time-discretized state trajectory with time discretization $\Delta t$,
\begin{equation}
    \V^{h,\pi}_T(\vx) \approx  \max_{0\leq k<H} h(\vx^\pi_{k\Delta t}) \coloneqq \V^{h,\pi}_H(\vx).
\end{equation}

This approximation may miss constraint violations between time steps. In \cite{knoedler2025safety}, a cubic spline is fitted to $h$ evaluated at the timesteps to estimate inter-step violations.

A robust extension of the \gls{pcbf} formulation, \gls{rpcbf}, considers disturbed dynamics in which the disturbance may enter the system through a general, potentially state- and input-dependent channel
\begin{equation}
\dot{\vx}=\tilde{F}(\vx,\vu,\vd), 
\end{equation}
with $\vd \in \cD$ where \(\mathcal{D}\) is defined by
\begin{equation}
\mathcal{D}
=
\left\{
\vd \in \mathbb{R}^{n_d}
\;\middle|\;
-\vd_m \preceq \vd \preceq \vd_m
\right\},
\qquad
\vd_m \in \mathbb{R}_{\geq 0}^{n_d},
\end{equation}
and \(\preceq\) denotes an elementwise inequality.

To approximate the robust finite-horizon value, $N$ disturbance realizations are sampled from $\mathcal{D}$ and the worst-case value over the resulting rollouts is used:
\begin{align}
\V^{h,\pi}_T(\vx_0)
&\approx
\V^{h,\pi}_{T,N}(\vx_0)
\coloneqq
\max_{i=1,\ldots,N}
\sup_{0\le t<T}
h(\vx_t^i),\\
\dot{\vx}_t^i
&=
\tilde{F}(\vx_t^i,\pi(\vx_t^i),\vd^i),
\qquad
\vd^i\sim\mathcal D,
\label{eq:pcbf}
\end{align}
where each $\vd^i$ denotes a sampled disturbance realization.

We approximate $\V^{h,\pi}_{T,N}$ by its time-discretized counterpart $\V^{h,\pi}_{H,N}$, for brevity, we omit its formal definition.
For improved readability, we omit the sub- and superscripts of $\V^{h,\pi}_{H, N}$ in the following and simply write $\V$ whenever the distinction is not required.

\subsection{ Approximation for Optimization via Linear Interpolation}\label{sec:pre_dev_free}

Gradient-based nonlinear optimization requires derivatives of the objective/constraints with respect to the optimization variables. For the \gls{cbf}-\gls{sf}, this includes the Jacobian of the nonlinear \gls{cbf} constraint with respect to the control input, which depends on the Jacobian of the system dynamics with respect to the control input. We next review a method for solving nonlinear optimization problems by constructing a local affine approximation from sampled function evaluations following~\cite{tracy2025trajectory}.

Consider the nonlinear optimization problem
\begin{equation}
\min_{\vz \in \mathbb{R}^{n_z}} \quad \|r(\vz)\|^2_2 \quad
\text{s.t.} \;  c(\vz) \leq 0,
\label{eq:nlp}
\end{equation}
where $r:\mathbb{R}^{n_z}\rightarrow\mathbb{R}^{n_r}$ and $c:\mathbb{R}^{n_z}\rightarrow\mathbb{R}^{n_c}$ are generally nonlinear residual and constraint functions, respectively.
Linear-interpolation-based optimization avoids requiring derivatives with respect to $\vz$ by constructing local affine approximations of the nonlinear functions from sampled evaluations. Given $M$ samples $\{\vz^1,\ldots,\vz^M\}$ around the current iterate $\bar{\vz}$, the sample points and function values are collected as
\begin{align}
\vW_z &=
\begin{bmatrix}
\vz^{1} & \vz^{2} & \cdots & \vz^{M}
\end{bmatrix}
\in\mathbb{R}^{n_z\times M},\\
\vW_r &=
\begin{bmatrix}
r(\vz^{1}) & r(\vz^{2}) & \cdots & r(\vz^{M})
\end{bmatrix}
\in\mathbb{R}^{n_r\times M},\\
\vW_c &=
\begin{bmatrix}
c(\vz^{1}) & c(\vz^{2}) & \cdots & c(\vz^{M})
\end{bmatrix}
\in\mathbb{R}^{n_c\times M}.
\end{align}
Any point within the convex hull of the sampled points can be expressed as
\begin{equation}
\vz = \vW_z\vlambda,
\end{equation}
where the interpolation weights satisfy
\begin{equation}
\Delta^{M-1}
=
\left\{
\vlambda\in\mathbb{R}^M
\;\middle|\;
\mathbf{1}^\top\vlambda=1,\;
\vlambda\ge0
\right\}.
\end{equation}

Replacing the residual/constraint functions in \eqref{eq:nlp} with their linear interpolation approximations yields the following relaxed  approximation of the original optimization problem:
\begin{equation}
\begin{aligned}
\min_{\vlambda,\vs}\quad
&\norm{\vW_r\vlambda}_2^2+\mu\mathbf{1}^\top\vs\\
\text{s.t.}\quad
&\vW_c\vlambda-\vs\leq 0, \;
\vs\geq 0, \;
\vlambda\in\Delta^{M-1}.
\end{aligned}
\label{eq:relaxed_qp}
\end{equation}
This is a convex QP in the interpolation weights $\vlambda$ and nonnegative slack vector $\vs\in\mathbb{R}^{n_c}$. It can be solved without derivatives of the original nonlinear objective or constraints with respect to $\vz$. Its solution $\vlambda^\star$ yields $\vz^\star=\vW_z\vlambda^\star$.

\subsection{Problem Formulation}
We assume access to a predictive model $\mathcal{M}$ of the robot-environment dynamics that generates forward rollouts of the robot-environment state $\vx$ under robot control inputs $\vu$. The model may be a physics-based simulator or learned \gls{wm}, and its derivatives with respect to the state or control input may not be available. Given a nominal input $\vu_\s{nom}$, we aim to construct a \gls{rpcbf}-\gls{sf} that minimally modifies the nominal control while avoiding $\avoidset$, using only forward evaluations of $\mathcal{M}$. The backup policy $\pi$ is treated as a design parameter, as its choice influences the conservativeness of the resulting \gls{sf}. The formulation should also remain applicable when $\mathcal{M}$ is updated or replaced, without requiring the \gls{cbf} to be retrained.

\section{Method}\label{sec:method}
We address the problem of \gls{cbf}-based safety filtering for tasks involving complex robot--environment interactions when derivatives of the predictive model are unavailable or unreliable. To this end, we develop a method for solving the \gls{sf} optimization using only forward-model evaluations, without requiring derivatives of the predictive model. To enable adaptation to changes in the predictive model, we employ implicit \glspl{cbf}, specifically \glspl{rpcbf}~\cite{knoedler2025safety}. Together, these components form Sim2CBF, our rollout-based framework for failure avoidance involving robot--environment interactions.

\begin{figure*}[t]
    \centering
    \includegraphics[width=0.8\linewidth]{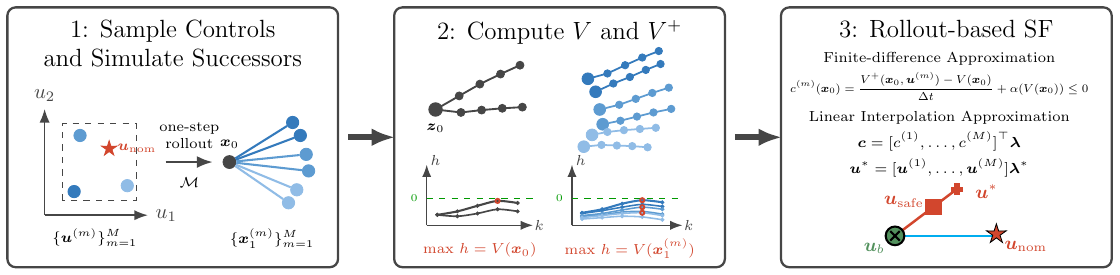}
    \caption{\textbf{Overview of the three components of the Sim2CBF Framework.}}
    \label{fig:overview}
    \vspace{-4mm}
\end{figure*}

Rather than blending directly between $\vu_\s{nom}$ and the backup control input $\vu_b$, as done in \cite{9691815}, we first compute a control input from an approximation of the original \gls{sf} optimization problem and then blend this control input with $\vu_b$. This requires substantially fewer control samples than a purely sampling-based \gls{sf}, yields less conservative behavior than direct nominal--backup blending, as we show in \cref{sec:results}, and retains failure avoidance guarantees through the blending mechanism. 
An overview of the proposed method is shown in \cref{fig:teaser}, with further details provided in \cref{fig:overview} and \cref{alg:der_free_pcbf}.

\subsection{Rollout-Based Safety Filter Optimization}
We begin by computing the control input that will subsequently be blended with $\vu_b$. 
When applied to \glspl{rpcbf}, the \gls{sf} in \eqref{eq:cbf_filter} requires derivatives of the dynamics at two stages.
Evaluating the \gls{cbf} constraint \eqref{eq:cbf_constraint} involves differentiating the \gls{rpcbf} with respect to the state, which requires differentiating through the finite-horizon policy rollout and underlying model. Solving the resulting optimization with gradient-based methods further requires derivatives of the constraint with respect to the control input.

To avoid computing the \gls{cbf} derivative, we approximate the Lie derivative in \eqref{eq:cbf_constraint} by
\begin{equation}\label{eq:finite_difference_approx}
\nabla\V(\vx)^\top F(\vx,\vu)
\approx
\frac{\V(\vx_{t+\Delta t})-\V(\vx_t)}{\Delta t},
\end{equation}
yielding
\begin{equation}\label{eq:finite_difference}
\frac{\V(\vx_{t+\Delta t})-\V(\vx_t)}{\Delta t}
\leq -\alpha\!\left(\V(\vx_t)\right).
\end{equation}
This condition is equivalent to a discrete-time \gls{cbf} condition with $\gamma(r)=\Delta t\alpha(r)$, ensuring failure avoidance at the prediction instants. It approximates the continuous-time condition as $\Delta t\to0$ and guarantees safety at $t+\Delta t$, but not necessarily throughout the interval $(t,t+\Delta t)$~\cite{tan2025zero}.

The finite-difference approximation~\eqref{eq:finite_difference} avoids explicitly computing the~\gls{cbf} derivative, but generally results in a constraint that is nonlinear in the control input $\vu$.
To solve the nonlinear \gls{sf} optimization problem efficiently without differentiating through the dynamics, we construct affine approximations of the objective and constraints by evaluating them at sampled control inputs and interpolating between these samples using convex combination weights.

At the current state $\vx_0=\vx$, consider $M$ candidate control inputs $\{\vu^{(m)}\}_{m=1}^M$, selected as described in \cref{sec:u}. Each candidate is applied for a single prediction step to obtain the successor state $\vx_1^{(m)}$.
Following the finite-horizon \gls{rpcbf} formulation, we then estimate $V(\vx_0)$ and $V(\vx_1^{(m)})$ using $N$ disturbance rollouts over a horizon $H$ under the backup policy $\pi$. This requires $(M+1)N$ finite-horizon rollouts and yields the \gls{cbf} value estimates used to construct the linear approximation of the \gls{sf} problem.

For each candidate control $\vu^{(m)}$, we evaluate the finite-difference \gls{cbf} constraint
\begin{equation}\label{eq:finite_diff_cond}
c^{(m)}
=
\frac{\V(\vx_1^{(m)})-\V(\vx_0)}{\Delta t}
+\alpha\left(\V(\vx_0)\right).
\end{equation}
Collecting these evaluations as
\begin{equation}
\vc=
\begin{bmatrix}
c^{(1)} & \cdots & c^{(M)}
\end{bmatrix}^{\top}.
\end{equation}
The \gls{cbf} constraint at $\vu=\vU^\top\vlambda$ is approximated by the linear interpolation $\vc^\top\vlambda$ with
\begin{equation}
\vU^\top =
\begin{bmatrix}
\vu^{(1)} & \cdots & \vu^{(M)}
\end{bmatrix}
\in \mathbb{R}^{n_u\times M},
\end{equation}
and $\vlambda\in\Delta^{M-1}$.
The \gls{sf} is then approximated as
\begin{equation}
\label{eq:qp}
\begin{aligned}
\vlambda^\star
=\argmin_{\vlambda\in\Delta^{M-1}}
\quad &
\norm{
\vU^\top\vlambda-\vu_\s{nom}
}^{2}
+\mu\mathbf{1}^\top\vs \\
\text{s.t.} \quad &
\vc^\top\vlambda -\vs \leq 0,
\quad \vs\geq 0.
\end{aligned}
\end{equation}
The filtered control input is recovered as
\begin{equation}
\vu^\star=\vU^\top\vlambda^\star.
\end{equation}
Since the objective is quadratic in $\vlambda$ and the interpolated \gls{cbf} constraint is linear in $\vlambda$, the resulting problem is a convex \gls{qp} with analytically available derivatives with respect to $\vlambda$, while avoiding differentiation through the system dynamics or rollout with respect to the control input.

\begin{algorithm}[t]
\caption{Rollout-based \gls{rpcbf}-\gls{sf}}
\label{alg:der_free_pcbf}
\footnotesize
\begin{algorithmic}[1]
\Require Current state $\vx_0$, backup policy $\pi$, failure function $h$, $\#$ prediction steps $H$, $\#$ disturbance samples $N$, $\#$ control candidate samples $M$
\Ensure Safe control input $\vu^\star$
\State Evaluate $V_{H,N}^{h, \pi}(\vx_0)$ 
\State Sample $M$ candidate control inputs
\For{$m=1,\ldots,M$}
    \State Predict $\vx_1^{(m)}$ using $\vu^{(m)}$
    \State Evaluate $V_{H,N}^{h, \pi}(\vx_1^{(m)})$
    \State Evaluate the CBF constraint $c(\vu^{(m)})$ \eqref{eq:finite_diff_cond}
\EndFor
\State Construct $\vU$ and $\vc$
\State Solve the interpolation-based quadratic program \eqref{eq:qp}
\State $\vu^\star=\vU^\top\vlambda^\star$
\State Compute $\vu_\s{safe}$ \eqref{eq:usafe}
\State \Return $\vu_\s{safe}$
\end{algorithmic}
\end{algorithm}

\subsection{Sampling Candidate Control Inputs}\label{sec:u}
For the optimization problem to admit a solution without constraint relaxation, at least one candidate control input must satisfy the \gls{cbf} constraint, i.e., $c^{(m)}\leq 0$ for some $m\in\{1,\ldots,M\}$.
One simple approach is to select candidate inputs independently along each control dimension using prescribed lower and upper bounds. However, this can result in a large spread of the candidates and, consequently, poor local linear approximations of the \gls{cbf} constraint. Instead, we include the nominal input $\vu_\s{nom}$ and the backup input $\vu_b$, which by construction satisfies the original \gls{cbf} constraint~\eqref{eq:cbf_constraint}, and augment them with candidate inputs corresponding to small perturbations in task space.

\subsection{Approximation Error and Failure Avoidance Guarantees}
We considered two approximations: the finite-difference approximation~\eqref{eq:finite_difference_approx} of the \gls{cbf} derivative and the linear interpolation of the resulting nonlinear constraint over the convex hull of the sampled control inputs.

The linear constraint in $\vlambda$ only approximates the nonlinear constraint over the convex hull of the sampled inputs. Hence, $\vc^\top\vlambda\leq0$ does not necessarily imply $c(\vU^\top\vlambda)\leq0$, and the interpolation error typically grows with the size of the sampled convex hull. The implication holds if $c(\vu)$ is convex over the sampled convex hull, and the approximation is exact if $c(\vu)$ is affine~\cite{vandenberghe2004convex}. 
A rigorous interpolation bound would require a Lipschitz constant for $c(\vu)$, which is difficult to obtain for black-box dynamics. This is unnecessary here, since failure avoidance follows from a separate argument.

We adopt the blending approach of~\cite{9691815} to establish the failure avoidance properties of the proposed controller. In contrast to the original formulation, we replace the nominal input $\vu_\s{nom}$ with the optimized input $\vu^\star$ obtained from the approximate \gls{sf} problem. The safe control input is therefore defined as
\begin{equation}\label{eq:usafe}
\vu_{\mathrm{safe}}
=
\beta(V)\vu^\star
+
\bigl(1-\beta(V)\bigr)\vu_b,
\qquad
\beta(V)\in[0,1],
\end{equation}
where $\beta(V)$ determines the interpolation between $\vu^*$ and the backup input $\vu_b$.
To recover the backup input on the boundary of the safe set, we impose
\begin{equation}
\beta(0)=0,
\end{equation}
such that
\begin{equation}
V=0
\quad\Longrightarrow\quad
\vu_{\mathrm{safe}}=\vu_b.
\end{equation}
Specifically, we choose $\beta(V)=1-\exp(\xi V)$, following~\cite{9691815}, where $\xi>0$ controls the blending rate. We further formalize the failure avoidance guarantee in the following.
\begin{lemma}
For an undisturbed system, let $\V^{h,\pi}_H(\vx)\leq0$. Then, the predicted trajectory under the backup policy $\pi$ avoids $\avoidset$ at each sampled time $k\Delta t$, $k=0,\ldots,H-1$.
\end{lemma}

\begin{proof}
If $V=0$, then by construction, the maximum constraint value encountered under the backup policy over the $H$ predicted discrete-time steps is zero. Hence, $h\leq 0$ at each predicted time step, and the failure set is avoided along the discretized rollout.
By definition,
$h(\vx^\pi_{k\Delta t})\leq\V^{h,\pi}_H(\vx)\leq0$
for all $k=0,\ldots,H-1$. Hence, none of the sampled states belongs to $\avoidset$.
\end{proof}

\begin{remark}
    Since the failure violation is evaluated only at discrete time steps, failure avoidance between consecutive samples is not guaranteed. As in~\cite{knoedler2025safety}, a cubic spline may be fitted to the sampled constraint values to better approximate violations between time steps. Moreover, in practice, the resulting safe control input $\vu_\s{safe}$ is typically applied in a zero-order-hold fashion between control updates, so continuous-time safety between updates likewise requires additional assumptions~\cite{tan2025zero}.
\end{remark}
We refer to our rollout-based implicit \gls{cbf}-\gls{sf} method as Interpolation and Blending.

\section{Results}\label{sec:results}
We evaluate our approach in experiments with increasing model complexity: first, for a low-dimensional simulated \gls{di}, then for simulated and hardware box-toppling with a 7-DoF manipulator, and finally for a Dubins car task with visual observations.

\begin{figure}
    \centering
    \includegraphics[width=0.8\linewidth]{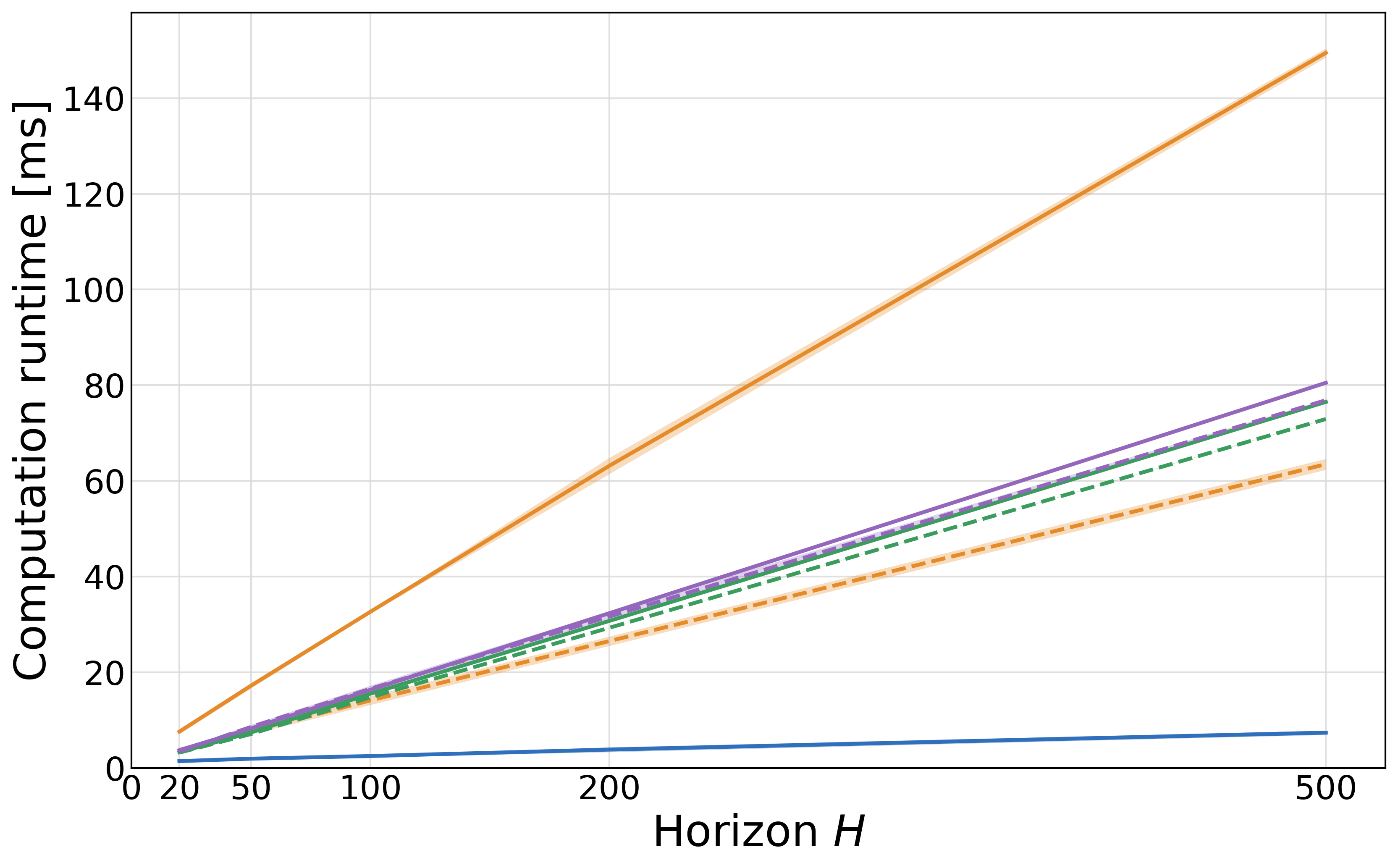}
    \includegraphics[width=0.8\linewidth]{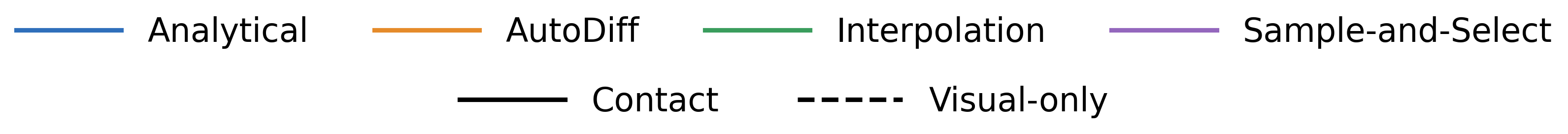}
    \caption{\textbf{Per-Timestep Computation Times for the 1D Double Integrator.}
    All methods use $N=50$. Interpolation uses $M=3$ candidate controls, requiring $(M+1)N=200$ MJWarp worlds, while Sample-and-Select uses $20$ grid samples over $\mathcal{U}$, requiring $1050$ worlds.}
    \vspace{-4mm}
    \label{fig:dbint_runtime}
\end{figure}

\subsection{Implementation Details}
Rollouts for the \glspl{rpcbf} are generated using two types of predictive models: physics-based simulators and learned \glspl{wm}. For physics-based simulation, we use MuJoCo~\cite{todorov2012mujoco} together with MuJoCo Warp~(MJWarp)~\cite{mujoco_mjwarp}, which enables GPU-accelerated parallel rollout generation and thus efficient evaluation. The proposed method is not specific to MuJoCo and can be applied with other predictive simulators/models. We further apply our approach in the latent space of a learned recurrent state-space \gls{wm} from RGB observations \cite{nakamura2025train}.

\begin{figure}
    \centering
    \includegraphics[width=0.75\linewidth]{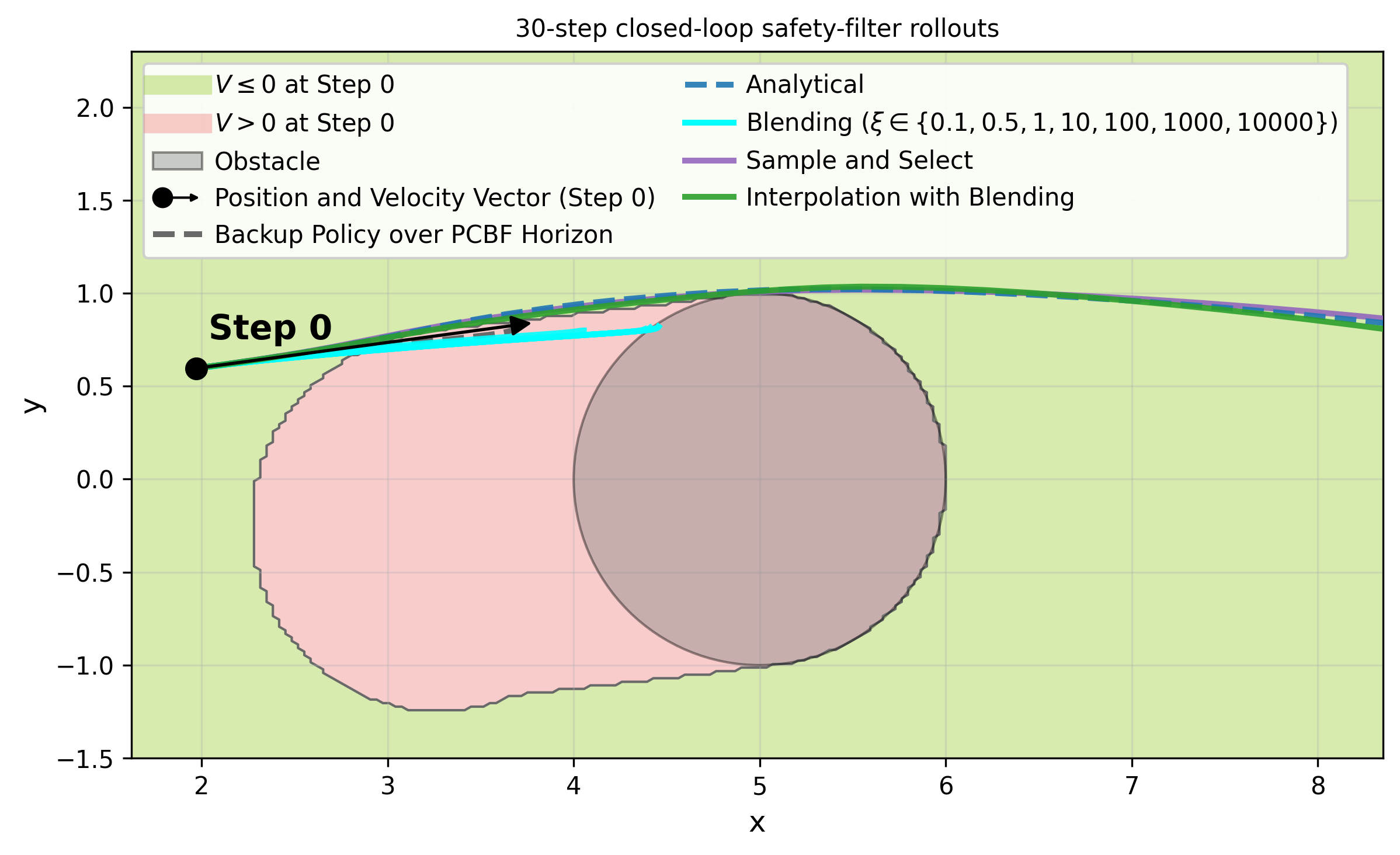}
    \caption{\textbf{Comparison of \glspl{sf} for 2D Double Integrator.}}
    \label{fig:2d_dbint}
       \vspace{-6mm}
\end{figure}

\subsection{Computation Time Comparison for the Double Integrator}
We first focus on simulator-based predictive models, which enable the framework to generalize across different settings by simply replacing the underlying robot and environment models in the simulator. This is particularly well suited to implicit \glspl{cbf}, as such changes can be accommodated without retraining the \gls{cbf}. 
We evaluate our approach on a 1D \gls{di}, i.e., $\dot{p}=v, \dot{v}=u$ with $\norm{u}\leq 1$ and constraints $\norm{p}\leq 1$. We compare three predictive models: analytical dynamics, MuJoCo~XLA~(MJX)\footnote{MuJoCo XLA provides JAX-based implementation and API, enabling hardware-accelerated MuJoCo simulation and automatic differentiation.}, and MJWarp.
Both the analytic dynamics and MJX allow the original \gls{sf} optimization problem \eqref{eq:cbf_filter} to be solved using exact (\textbf{Analytical}) or automatically differentiated (\textbf{AutoDiff}) model derivatives, respectively. In contrast, automatic differentiation is currently not available for MJWarp; therefore, we apply our proposed rollout-based \gls{sf} (\textbf{Interpolation}). Additionally, we evaluate \textbf{Sample and Select}, which evaluates the \gls{cbf} condition \eqref{eq:finite_difference} on sampled control inputs and selects a feasible candidate as an alternative to solving the \gls{sf} optimization with gradient-based methods. 
For the simulator-based models, we consider settings both with and without simulated contact dynamics. In the contact setting, the walls and floor participate in collision detection and generate contact forces; in the visual-only setting, they are rendered but excluded from collision handling. In all of the above cases, maximum braking is used as the backup policy. The computation times for experiments conducted on an NVIDIA GeForce RTX 4060 GPU are reported in \cref{fig:dbint_runtime}.
The experiments show that automatic differentiation becomes substantially slower when contact dynamics are simulated.
Given MJWarp's lower computation time when modeling contacts, we use MJWarp for all subsequent simulator-based predictive models.
Next, we compare the resulting trajectories when applying the different \glspl{sf} to an undisturbed 2D Double Integrator collision avoidance task. 
The results can be seen in \cref{fig:2d_dbint}. While Analytical, Sample-and-Select, and Interpolation with Blending successfully avoid collisions and progress toward the goal, Blending alone becomes unsafe and gets stuck behind the obstacle. This behavior may result from the conservatism of reverting toward the backup policy, which can limit progress, while the zero-order-hold implementation of the control input may lead to safety violations between control updates.

\subsection{Simulation Experiments for Box Toppling Task}
We conduct simulation experiments on a joint-torque-controlled 7-DoF Franka Panda manipulator performing a box-toppling task. The box mass is assumed unknown but bounded within $[1.3,1.7]$ Kg. We account for this uncertainty when constructing the RPCBF by considering rollouts at the minimum and maximum admissible masses. Interpolation and Blending achieves a safety rate of $0.92$ over $50$ trials.

\begin{figure*}[!t]
    \centering
    \includegraphics[width=0.83\linewidth]{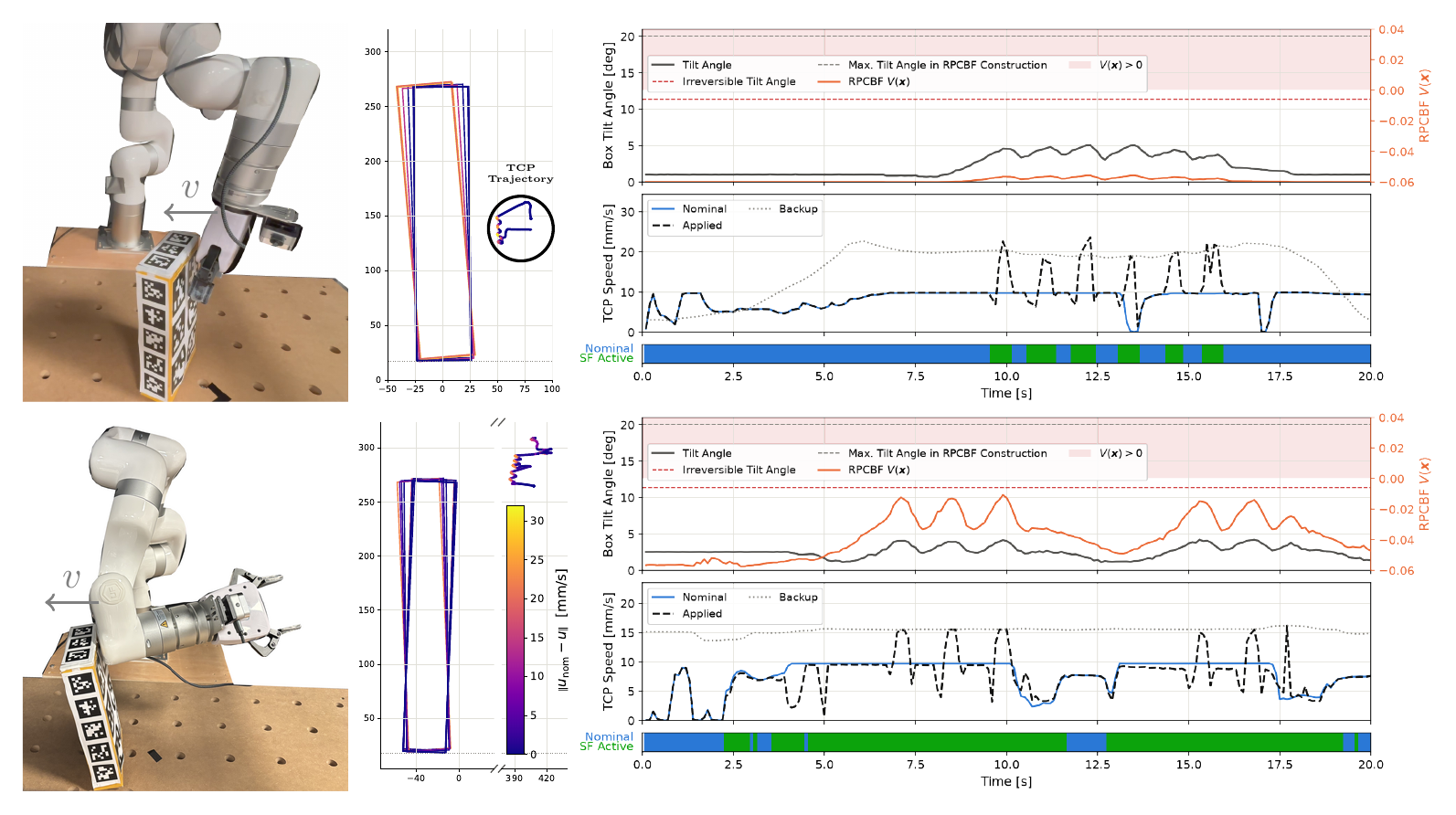}
    \caption{\textbf{Hardware experiments for the cereal-box toppling task on the xArm 7.} Left: experimental setup. Middle: side view of the box illustrating the \gls{sf} correction of the nominal control $\vu_\s{nom}$ at different box tilt angles. Right: box tilt angle, \gls{rpcbf} value, nominal control, backup control, and applied safe control over time.}
    \label{fig:hardware}
       \vspace{-4mm}
\end{figure*}

\subsection{Hardware Experiments for Box Toppling Task}
We conduct hardware experiments on a velocity-controlled UFACTORY xArm 7, performing a cereal-box toppling task to assess the feasibility of our approach on a physical system. The box pose is estimated using AprilTags, while its dimensions and weight are assumed to be known. We apply the Sim2CBF framework, using the interpolation and blending approach to solve the \gls{sf} optimization problem with $N=50$, $M=9$, $H=100$. The backup policy retracts the robot toward its initial configuration, while the nominal policy is a teleoperation policy that pushes the box to induce toppling. Mean and maximum solve time of the \gls{sf} are $45.02\mathrm{ms}$ and $66.41\mathrm{ms}$, respectively.
The nominal \gls{tcp} speed, \gls{sf} corrections, \gls{rpcbf} value, and resulting box tilt angle are shown in~\cref{fig:hardware} for two representative trials. Note that the function $h$ used to construct the \gls{rpcbf} does not explicitly encode the irreversible tilt angle. Instead, the maximum allowable tilt angle is set beyond this threshold, and the \gls{rpcbf} determines through its rollouts when corrective action is required to avoid irreversible toppling. Due to the perception errors observed during the experiments and accounted for in the rollouts, the resulting \gls{rpcbf} is conservative; improved perception would therefore reduce this conservatism.
We highlight an advantage of using simulators as predictive models: contacts involving any part of the robot are accounted for, as demonstrated by the ``elbow''-pushing experiment in the bottom row  of~\cref{fig:hardware}. This contrasts with many vision-based \gls{wm} approaches, which often focus primarily on observations of the end-effector and its local interactions with the environment.

\begin{figure}[!t]
\centering
    \includegraphics[width=0.87\linewidth]{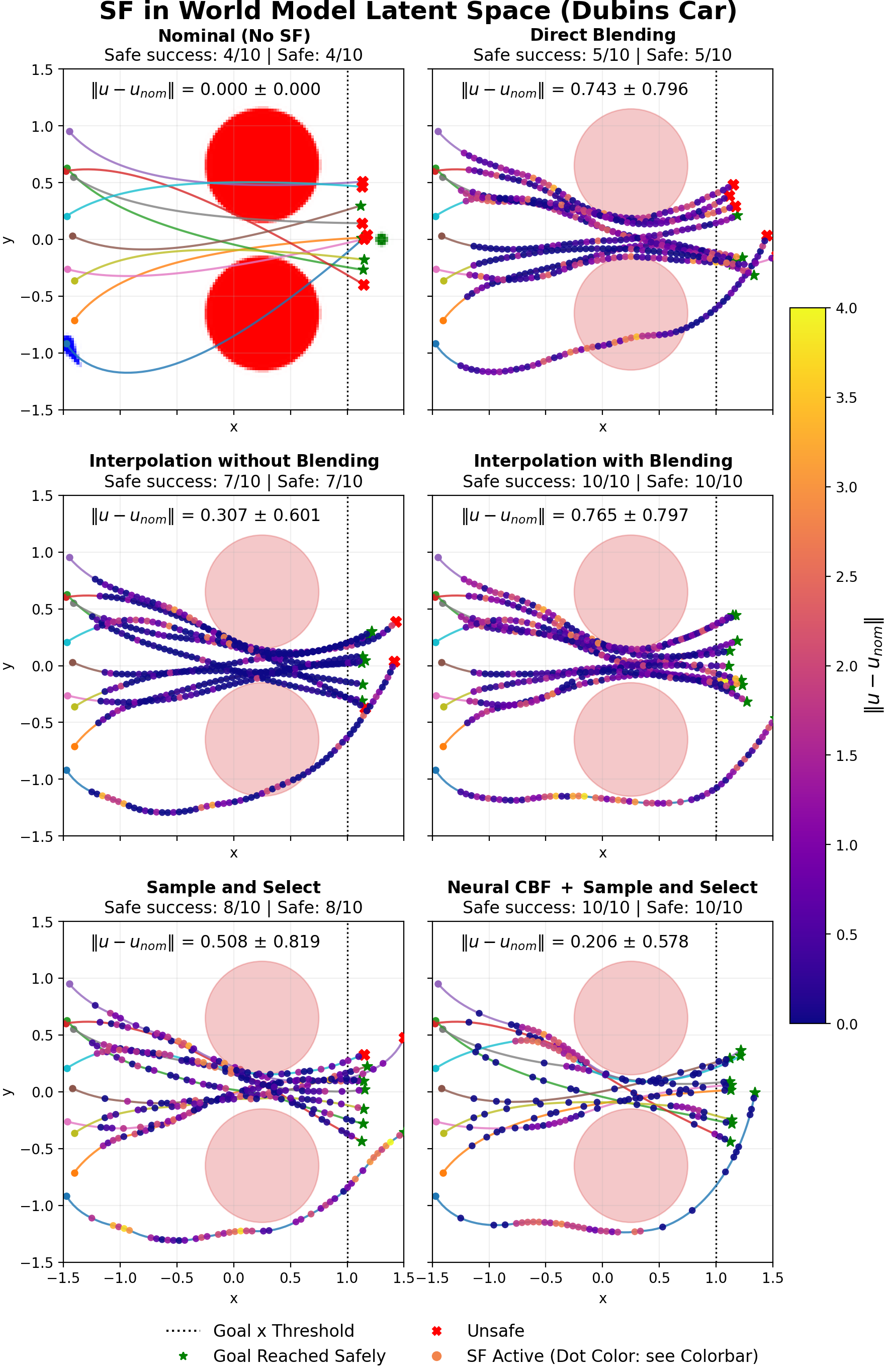}
\caption{\textbf{\gls{sf} Comparison for Dubins' Car in \gls{wm} Latent Space.} Example adapted from \cite{nakamura2025train}. The first-panel background shows the \gls{wm} input, while each panel shows the resulting trajectory for the corresponding \gls{sf}. Dot color indicates the control deviation $\norm{\vu-\vu_\s{nom}}$.}
    \label{fig:latent_traj}
    \vspace{-7mm}
\end{figure}

\subsection{Dubins Car in World Model Latent Space}
Finally, we apply rollout-based \glspl{sf} in the latent space of a learned \gls{wm} on a Dubins car collision-avoidance task, following \cite{nakamura2025train}. We evaluate our full method, \textbf{Interpolation with Blending}, an \textbf{Interpolation without Blending} ablation, and compare both against \textbf{Direct Blending}~\cite{9691815} and \textbf{Sample and Select}~\cite{nakamura2025train}. All use \gls{rpcbf} as the underlying \gls{cbf}, with a greedy backup policy minimizing the worst-case next-step $h$. We also consider a \textbf{Neural \gls{cbf} + Sample and Select} baseline~\cite{nakamura2025train}. \Cref{fig:latent_traj} shows ten trajectories generated from randomly sampled initial conditions and goals. The nominal policy is a proportional-derivative heading controller. We report the safe success rate, defined as the fraction of rollouts that avoid the obstacle while reaching beyond $x=1$. \Cref{tab:latent_2obst} provides extended results for $100$ trials.

\begin{table}[t]
    \centering
    \caption{\textbf{\gls{sf} Comparison for Dubins Car in WM Latent Space over 100 Initial--Goal Pairs.} \textnormal{Safety rate is omitted as it equals safe success for all methods. For $\norm{\vu-\vu_\s{nom}}$, mean and standard deviation are reported.}}
    \label{tab:latent_2obst}
    \scriptsize
    \setlength{\tabcolsep}{3pt}
    \begin{tabular}{lccc}
        \hline
        \textbf{Method} 
        & \shortstack{\textbf{Safe Success Rate} $\uparrow$}
        & \shortstack{$\norm{\vu - \vu_\s{nom}}$$\downarrow$} \\
        \hline
        Nominal (No SF)
        & 0.45 & 0.00 $\pm$ 0.00 \\
        
        Direct Blending
        & 0.86 & 0.50 $\pm$ 0.80 \\
        
        Interpolation without Blending (Ours)
        & 0.78  &  0.27 $\pm$ 0.52 \\

        Interpolation with Blending (Ours)
        & 0.95 & 0.67 $\pm$ 0.79 \\

        Sample and Select
        & 0.82 & 0.49 $\pm$ 0.75 \\

        Neural CBF + Sample and Select
        & 1.00 & 0.17 $\pm$ 0.52 \\
        \hline
    \end{tabular}
    \vspace{-6mm}
\end{table}

The neural \gls{cbf} achieves the highest safe success rate, which is likely due to stochastic sampling of the \gls{wm} during training, whereas our method accounts for uncertainty using only $N=50$ rollouts of the stochastic \gls{wm}. It also yields the smallest average correction to $\vu_{\mathrm{nom}}$, likely because the diffusion-based backup policy used during training already produces comparatively safe actions. In contrast, the backup policy of \gls{rpcbf} is less capable and therefore yields a more conservative \gls{sf}, correcting $\vu_{\mathrm{nom}}$ earlier.
We demonstrate in \cref{tab:latent_change} that \gls{rpcbf} readily adapts to updated \glspl{wm} or safety constraints, while neural \glspl{cbf} generally require retraining. Overall, Interpolation with Blending provides flexible failure avoidance under changing models.

\begin{table}[t]
\centering
\caption{\textbf{Adaptability of Neural and Implicit \glspl{cbf} as Model changes.} \textnormal{Neural \gls{cbf} is trained on the two-obstacle environment. For both \glspl{cbf}, the Safe Success Rate $\uparrow$/ Safe Rate $\uparrow$ are reported after retraining the \gls{wm} and $h$ (MF). The one obstacle is centered at $y=0$.}}
\label{tab:latent_change}
\scriptsize
\setlength{\tabcolsep}{3pt}
\begin{tabular}{lcc}
\hline
\noalign{\vskip 3pt}
\diagbox[height=0.7cm]{\textbf{Method}}
{\raisebox{-1.9mm}[0pt][0pt]{%
    \hspace{3mm}%
    \shortstack{\textbf{Obstacle}\\\textbf{Retrained}}}}
& \shortstack{\textbf{2 Obs.}\\ \textbf{(WM + MF)}}
& \shortstack{\textbf{1 Obs.}\\ \textbf{(WM + MF)}}\\ 
\noalign{\vskip 3pt}
\hline

Interpolation with Blending
& 0.95/0.95 &  0.94/0.94 \\

Sample and Select
& 0.82/0.82 &0.94/0.96 \\

Neural \gls{cbf} + Sample and Select
& 1.00/1.00  & 0.11/0.11 \\

\hline
\end{tabular}
   \vspace{-6mm}
\end{table}

\section{CONCLUSION}
\glsresetall
We presented Sim2CBF, an adaptable rollout-based failure-avoidance framework for settings involving complex robot–environment interactions. By combining finite-difference approximations with local interpolation over sampled controls, Sim2CBF approximates the original \gls{cbf}-\gls{sf} as a \glslong{qp} using only forward evaluations of the predictive model, without differentiating through the dynamics. This enables implicit \glspl{cbf}, evaluated through model rollouts rather than available in closed form, to be used with non-differentiable predictive models or models whose derivatives are unreliable, including contact-rich simulators and learned \glspl{wm}. Experiments, including box-toppling tasks, demonstrate effective failure avoidance.

Future work includes online model adaptation to better exploit implicit \gls{cbf} adaptability, evaluation in richer simulations with more complex and semantically meaningful failures, and studying failure avoidance between discrete control updates under zero-order hold.

\section{Acknowledgement}
We thank Florian T. Pokorny for his support in enabling the hardware experiments.









\bibliographystyle{IEEEtran}
\bibliography{references}

\end{document}